\documentclass{llncs}
\usepackage{makeidx}
\usepackage{amssymb}
\usepackage{amsmath}
\usepackage{amsfonts}
\usepackage{epstopdf}
\usepackage{multirow}
\usepackage{graphicx}
\usepackage{caption}
\usepackage{enumerate}
\usepackage{pgfplots}
\usepackage{bm,graphicx}
\usepackage{booktabs}
\usepackage{algorithm}% http://ctan.org/pkg/algorithm
\usepackage{algpseudocode}% http://ctan.org/pkg/algorithmicx
\usepackage{subfigure}
\begin{document}

\title{Brain Biomarker Interpretation in ASD Using Deep Learning and fMRI}

%\titlerunning{Lecture Notes in Computer Science: Authors' Instructions}

%\author{Xiaoxiao Li\and James Duncan}
\author{Xiaoxiao Li$^{\star}$, Nicha C. Dvornek$^{\dagger}$,  Juntang Zhuang$^{\star}$, Pamela Ventola$^{\ddagger}$ and \\ James S. Duncan$^{\star\ast\dagger}$}

\institute{Department of Biomedical Engineering, Yale University,\\ New Haven, CT 06520, USA}

\institute{$^{\star}$ Biomedical Engineering, Yale University, New Haven, CT USA \\
	$^{\ast}$ Electrical Engineering, Yale University, New Haven, CT USA\\
	$^{\dagger}$Radiology \& Biomedical Imaging, Yale School of Medicine, New Haven, CT USA\\
	$^{\ddagger}$ Child Study Center, Yale School of Medicine, New Haven, CT USA}
%\author{Xiaoxiao Li, Nicha C. Dvornek,  Juntang Zhuang, Pamela Ventola and \\ James S. Duncan}
%\institute{Yale University, New Haven, CT USA}
\maketitle
\footnotetext[1]{This work was supported by NIH Grant  5R01 NS035193.}

\begin{abstract}
Autism spectrum disorder (ASD) is a complex neurodevelopmental disorder. Finding the biomarkers associated with ASD is extremely helpful to understand the underlying roots of the disorder and can lead to earlier diagnosis and more targeted treatment. Although Deep Neural Networks (DNNs) have been applied in functional magnetic resonance imaging (fMRI) to identify ASD, understanding the data driven computational decision making procedure has not been previously explored. Therefore, in this work, we address the problem of interpreting reliable biomarkers associated with identifying ASD; specifically, we propose a 2-stage method that classifies ASD and control subjects using fMRI images and interprets the saliency features activated by the classifier. First, we trained an accurate DNN classifier. Then, for detecting the biomarkers, different from the DNN visualization works in computer vision, we take advantage of the anatomical structure of brain fMRI and develop a frequency-normalized sampling method to corrupt images. Furthermore, in the ASD vs. control subjects classification scenario, we provide a new approach to detect and characterize important brain features into three categories. The biomarkers we found by the proposed method are robust and consistent with previous findings in the literature. We also validate the detected biomarkers by neurological function decoding and comparing with the DNN activation maps. 
\end{abstract}

\section{Introduction}

Autism spectrum disorder (ASD) affects the structure and function of the brain. To better target the underlying roots of ASD for diagnosis and treatment, efforts to identify reliable biomarkers are growing \cite{goldani2014biomarkers}. Significant progress has been made using functional magnetic resonance imaging (fMRI) to characterize the brain changes that occur in ASD \cite{Kaiser07122010}.  

%Recently, many deep neural networks have done effectively at classification ASD using fMRI. Ididaka et al. \cite{iidaka2015resting} used pixel neural network, Dvornek et al. \cite{dvornek2017identifying} used long short-term memory (LSTM) network and Parisot et al. used graph convolutional network \cite{parisot2017spectral},Li at al use 3D convolutional networks \cite{XiaoliISBI} on the ASD fMRI classification. 
Recently, many deep neural networks (DNNs) have been effective at identifying ASD using fMRI \cite{iidaka2015resting,XiaoliISBI}. However, these methods lack model transparency. Despite promising results, the clinicians typically want to know if the model is trustable and how to interpret the results. Motivated by this, here we focus on developing the interpretation method for deciphering the regions in fMRI brain images that can distinguish ASD vs. control by the deep neural networks. 

There are three main approaches for interpreting the important features detected by DNNs. One approach is using gradient ascent methods to generate an image that best represents the class \cite{yosinski2015understanding}. However, this method cannot handle nonlinear DNNs well. The second approach is to visualize how the network responds to a specific corrupted input image in order to explain a particular classification made by the network \cite{zintgraf2017visualizing}. The third one uses the intermediate outputs of the network to visualize the feature patterns \cite{zhou2016learning}. However, all of these  existing methods tend to end up with blurred and imprecise saliency maps. 

%The goal of our work is to identity biomarkers for ASD, defined as important regions of interest (ROIs) in the brain that distinguish autistic and healthy controls. We propose a method that belongs to the second feature detection approach. Different from the above methods, we utilize the structure of the brain. The  ROI-based method allows us to simplify the corrupted image generating procedure, and we analyze the features using the distribution of DNN predictions and statistical hypothesis testing. In addition, we validate our findings with DNN intermediate outputs.

The goal of our work is to identity biomarkers for ASD, defined as important regions of interest (ROIs) in the brain that distinguish autistic and healthy controls. Different from traditional brain biomarker detection methods, by utilizing the high dimensional feature capturing ability of DNNs and brain structure, we propose an innovative 2-stage pipeline to interpret biomarkers. Different from above DNN visualization methods, our main contribution includes a ROI-based image corruption and generating procedure. In addition, we analyze the feature importance using the distribution of DNN predictions and statistical hypothesis testing. We applied the proposed method on multiple datasets and validated our robust findings by decoding neurological function of biomarkers, viewing DNN intermediate outputs and comparing literature reports.

%Generally speaking, our proposed method belongs to the second approach. Different from the above methods, we utilize the anatomical structure of brain; and considering our goal - characterize important regions of interest (ROIs) into different categories rather than assign an importance probability, our method simplified the corrupted image generating procedure and analyzes the features from probabilistic prediction distributions. Aside from this, we validate our findings with DNN intermediate outputs.  

%In this paper, we focus on the need of understanding DNN model in neuroimage analysis. Our contributions are 1) described a two-stage pipe line to interpret ASD brain features; 2) used transfer learning method and achieve high accuracy in classifying ASD vs. control; 3) proposed an image corruption scheme and a prediction probability calculation method for registered corrupted images; 4) developed a statistic analysis method to evaluate model prediction difference; 5) demonstrate the effectiveness of our procedure on different datasets; and 6) validated the interpretation results with previous literature, activation map and Neurosynth decoding results. 
   
%Except for the data driven model mentioned above, another approach to investigate the data distribution and its features is to mimic the observation data generating procedure. Venkataraman \cite{venkataraman2016bayesian} proposed an univariate generative model to detect abnormal community in ASDs. This work interpreted hyper-active and hypo active regions and networks from fMRI correlation matrix. 

\section{Method}
\subsection{Two-stage Pipeline With Deep Neural Network Classifier}
\begin{figure}[t]
	\centering
	\centerline{\includegraphics[width=12cm]{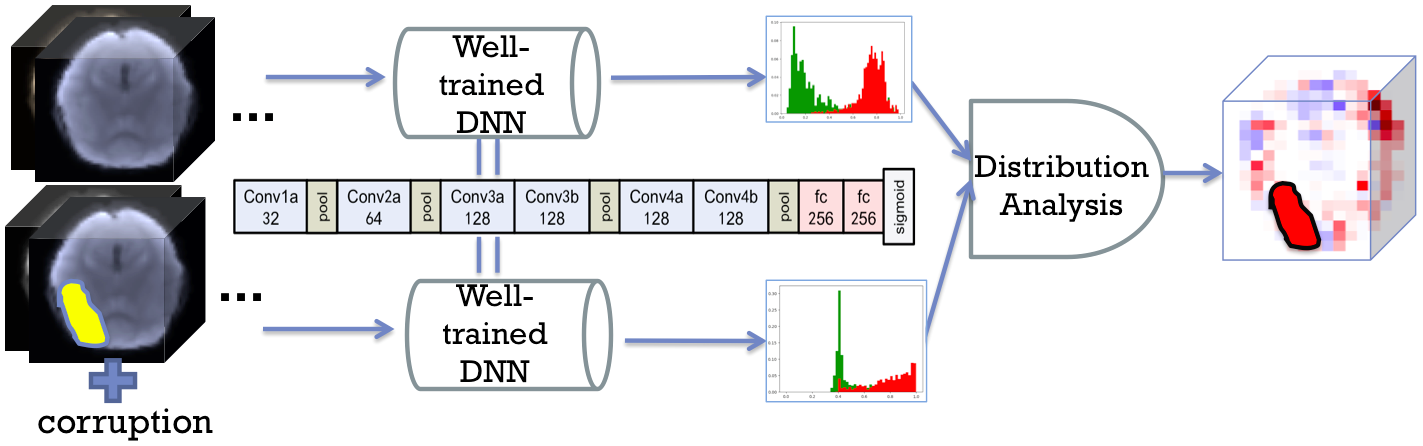}}
	%  \vspace{2.0cm}
	%\centerline{(a) Result 1}\medskip
	\caption{Pipeline for interpreting important features from a DNN}
	\label{diagram}
\end{figure}
We propose a corrupting strategy  to find the important regions activated by a well-trained ASD classifier (Fig. \ref{diagram}). The first stage is to train a DNN classifier for classifying ASD vs. control subjects. The DNN we use (2CC3D) has 6 convolutional, 4 max-pooling and 2 fully connected layers, followed by a sigmoid output layer \cite{XiaoliISBI} as shown in the middle of Fig. \ref{diagram}. The number of kernels are denoted on each layer in Fig. \ref{diagram}. Dropout and l2 regularization are applied to avoid overfitting. The study in \cite{XiaoliISBI} demonstrated that we can achieve higher accuracy using the 2CC3D framework, since it integrates spatial-temporal information of 4D fMRI. Each frame of 3D fMRI is downsampled to 32x32x32. We use sliding-windows with size $w$ and stride length $stride$ to move along the time dimension of the 4D fMRI sequence and calculate the mean and standard deviation (std) for each voxel's time series within the sliding window. 
 Given $T$ frames in each 4D fMRI sequence, by this method,  $\lfloor\frac{T-w}{stride}\rfloor+1$ 2-channel images (mean and std fMRI images) are generated for each subject. We define the original fMRI sequence as $I(x,y,z,t)$, the mean-channel sequence as $\tilde{I}(x,y,z,t)$ and the std-channel as $\hat{I}(x,y,z,t)$. For any $x,y,z$ in $\{0,1,\cdots,31\}$,
%{\setlength\abovedisplayskip{1pt plus 1pt minus 1pt} 
%	\setlength\belowdisplayskip{1pt plus 1pt minus 1pt} 
	\begin{equation}
	\tilde{I}(x,y,z,t)=\frac{\sum_{\tau=t+1-w}^{t}I(x,y,z,\tau)}{w}
	\end{equation}
	\begin{equation}
	\hat{I}(x,\!y,\!z,\!t)\!=\!\sqrt{\frac{\sum_{\tau=t+1-w}^{t}[I(x,\!y,\!z,\!\tau)\!-\!\tilde{I}(x,\!y,\!z,\!t)]^2}{w-1}}.
	\end{equation}
The outputs are probabilistic predictions ranging in $[0,1]$. The second stage is to interpret the output differences after corrupting the image. We corrupt a ROI of the original image and put it in the well-trained DNN classifier to get a new prediction (section 2.2). Based on the prediction difference, we use a statistical method to interpret the importance of the ROI (section 2.3).

\subsection{Prediction Difference Analysis}

\begin{algorithm}[t]
	\caption{Important Feature Detection For Binary Classification}\label{euclid}
	\hspace*{\algorithmicindent} \textbf{Input:} {$X^0$, a group of images from class 0; ${X^1}$, a group of images from class 1; $f$, DNN classification model.}
	%\hspace*{\algorithmicindent} \textbf{Output}
	\begin{algorithmic}[1]
		\State
		\texttt{$\mathbb{P}_{o}^0 \gets  f(X^0)$ and $\mathbb{P}_{o}^1\gets f(X^1)$ }
		\State
		\texttt{$JSD^{o}_{+/-} \gets JSD(\mathbb{P}_o^{0},\mathbb{P}_o^{1}) $}
		\Comment{by bootstrapping}
		\For{\texttt{$r$ in ROIs}}
		\State
		\texttt{$\mathbb{P}_{c}^0\gets f(X^0_{\setminus r} )$, $\mathbb{P}_{c}^1\gets f(X^1_{\setminus r})$ }
		\Comment{by sampling} 
		\State
		\texttt{$JSD^c_{+/-} \gets JSD(\mathbb{P}_c^0,\mathbb{P}_c^1)$, $Shift^0\gets \mathbb{P}^{0}_c-\mathbb{P}^{0}_{o}$, $Shift^1 \gets \mathbb{P}^1_c-\mathbb{P}^1_0$}     
		\If {$JSD^c_+<JSD^o_-$ or $median(\mathbb{P}_c^0) >median(\mathbb{P}_c^1)$}
		\Comment{fool the classifier}
		\State {\bf{do} {Wilcoxon(Shift) one tailed test}} 
		
		\If{$\mathbb{P}^0 \Rightarrow 1$ and $\mathbb{P}^1 \Rightarrow 0$}
		\State {$\bf{r}$ is an important feature for both classes}
		\ElsIf{only $\mathbb{P}^0 \Rightarrow 1$ }
		\State{$\bf{r}$ is an important feature for class 0}
		\ElsIf{only $\mathbb{P}^1 \Rightarrow 0$ }
		\State{$\bf{r}$ is an important feature for class 1}
		\EndIf
		\Comment{$ \Rightarrow$ means significant shift}
		\EndIf
		\EndFor		
	\end{algorithmic}
	
\end{algorithm}
We use a heuristic method to estimate the feature (an image ROI) importance by analyzing the probability of the correct class predicted by the corrupted image. 

%Simply speaking, we corrupt a ROI for all the images in the same class, and then analyze the shift of prediction outcome.  The basic idea is that the relevance of ROI can be estimated by measuring how the prediction changes if the feature is corrupted.

In the DNN classifier case, the probability of the abnormal class $c$ given the original image $\bm{X}$ is estimated from the predictive score of the DNN model $f$ : $ f(\bm{X}) = p(c|\bm{X})$. Denote the image corrupted at ROI $\bm{r}$ as $\bm{X_{\setminus r}}$. The prediction of the corrupted image is $p(c|\bm{X_{\setminus r}})$. To calculate $p(c|\bm{X_{\setminus r}})$, we need to marginalize out the corrupted ROI $\bm{r}$:
\begin{equation}
p(c|\bm{X_{\setminus r}}) = \mathbb{E}_{\bm{x_r}\sim p(\bm{x_r}|\bm{X_{\setminus r}})}p(c|\bm{X_{\setminus r}},\bm{x_r}),
\end{equation}
where $\bm{x_r}$ is a sample of ROI $r$. Modeling $p(\bm{x_r}|\bm{X_{\setminus r}})$ by a generative model can be computationally intensive and may not be feasible. We assumed that an important ROI contains features that cannot be easily sampled from  the same ROI of other classes and is predictive for predicting its own class. Hence, we approximated $p(\bm{x_r}|\bm{X_{\setminus r}})$ by sampling $\bm{x_r}$ from each ROI $\bm{r}$ in the whole sample set. In fMRI study, each brain can be registered to the same atlas, so the same ROI in different images have the same spatial location and number of voxels. Therefore, we can directly sample $\bm{{\hat{x}}_r}$s and replace $\bm{x_r}$ with them. Then we flatten the  $\bm{{\hat{x}}_r}$ and $\bm{x_r}$ as vectors $\overrightarrow{\bm{{\hat{x}}_r}}$ and $\overrightarrow{\bm{x_r}}$. From the $K$ sampled $\bm{{\hat{x}}_r}^k$s, we calculate the Pearson correlation coefficient $\rho_k = {cov(\overrightarrow{\bm{{\hat{x}^k_r}}},\overrightarrow{\bm{x_r}})}/{\sigma_{\overrightarrow{\bm{{\hat{x}^k_r}}}}\sigma_{\overrightarrow{\bm{x_r}}}}$, where $ k \in \{1,2,\dots,K\}$, $\rho \in [-1,1]$. Because sample size of each class may be biased, we will de-emphasize the samples that can be easily sampled, since $p(c|\bm{X_{\setminus r}})$ should be irrelevant to the sample set. Therefore, we will do a frequency-normalized transformation. We divide [-1,1] into $N$ equal-length intervals. Each $\rho_k$ will fall in one of the intervals. After $K$ samplings, we calculate $N_i$, the number of sample correlations in interval i, where $i \in \{1,2,\dots,N\}$. For the $\rho_k$ located in interval $i$, the frequency-normalized weight is $w_k = \frac{1}{N\cdot N_i}$.  Denote $\bm{X_k'}$ as the image $\bm{X}$ replacing $\bm{x_r}$ with $\bm{{\hat{x}}_r}^k$. Hence, we approximate $p(c|\bm{X_{\setminus r}})$ as
\begin{equation}
p(c|\bm{X_{\setminus r}})\approx \sum_k{w_k}p(c|\bm{X_k'}). 
\end{equation} 
% The difference of prediction is measured by $\Delta = p(c|\bm{X})$ - $p(c|\bm{X}_{\setminus r})$.
%\vspace{-.5cm}

\subsection{Important Feature Interpretation}
In the binary classification scenario, we label the reference class as 0 and the experiment class as 1. The original prediction probability of the two classes are denoted as $\mathbb{P}^0_o$ and $\mathbb{P}^1_o$, which are two vectors containing the prediction results $p(c|\bm{X})$s for each sample in the two classes respectively. Similarly, we have $\mathbb{P}^0_c$ and $\mathbb{P}^1_c$ containing $p(c|\bm{X}_{\setminus r})$s for the corrupted images. We assume that corrupting an important feature will make the classifier perform worse. One extreme case is that the two distributions shift across each other, which can be approximately measured by $median(\mathbb{P}_c^0) >median(\mathbb{P}_c^1)$. If this is not the case, we use Jensen-Shannon Divergence ($\bm{JSD}$) to measure the distance of two distributions:
\begin{equation}
JSD(\mathbb{P}_0,\mathbb{P}_1) = \frac{1}{2}KL(\mathbb{P}_0\parallel\frac{\mathbb{P}_0+\mathbb{P}_1}{2})+\frac{1}{2}KL(\mathbb{P}_1\parallel\frac{\mathbb{P}_0+\mathbb{P}_1}{2})
\end{equation}
where $KL(\mathbb{P}_0\parallel \mathbb{P}_1) = -\sum_i \mathbb{P}_0(i)log(\mathbb{P}_1(i)/\mathbb{P}_0(i))$. Given two distributions $\mathbb{P}_0$ and $\mathbb{P}_1$, we use bootstrap method to calculate the upper bound $JSD_+$ and the lower bound $JSD_-$ with confidence level, $1-\alpha_{JSD}$. We classify the important ROIs into different categories based on the shift of the prediction distribution before and after corruption. The one-tailed Wilcoxon paired difference test \cite{whitley2002statistics} is applied to investigate whether the shift is significant. We use false discovery rate (FDR) controlling procedure to handle testing the large number of ROIs. FDR adjusted q-value is used to compare with the significance level $\alpha_W$. The method to evaluate the feature importance is shown in Algorithm 1.
%\begin{figure}[htbp]
%	\centering
%	\centerline{\includegraphics[width=12cm]{shift}}
%	%  \vspace{2.0cm}
%	%\centerline{(a) Result 1}\medskip
%	\caption{Simple diagram of the interpretation criterion}
%	\label{shift}
%	%
%\end{figure}

\section{Experiments and Results}

\subsection{Experiment 1: Synthetic Data Model}
\begin{figure}[t]
	\begin{minipage}[]{.5\linewidth}
		\centering
		\includegraphics[width = 4cm]{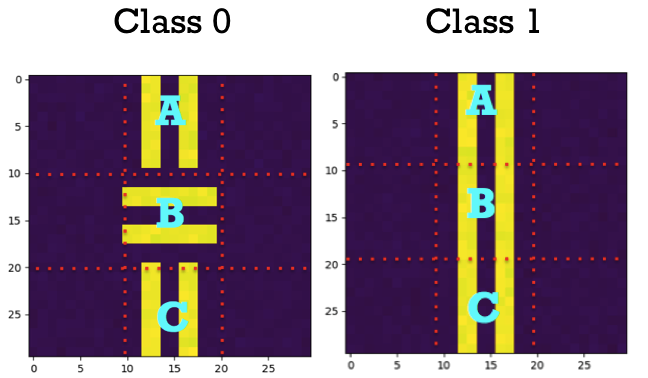}
		
		\caption{Synthetic Images }
		
		\label{synth}
	\end{minipage}
	\begin{minipage}[]{.5\linewidth}
		\centering
		\begin{tabular}{c|c|c}
			\hline
			& \textbf{Class 0} & \textbf{Class 1} \\\hline
			\textbf{Equal}&$0.10\!\pm\!0.01$& $0.91\!\pm\!0.03$ \\ \hline
			\textbf{Normalize}&$0.49\!\pm\!0.02$&$0.50\!\pm\!0.01$\\ \hline
		\end{tabular}
		\label{tab}
		\centering
		\captionsetup{justification = centering}
		\captionof{table}{Misclassfication rate when corrupting patch B}	
	\end{minipage}
\end{figure}

We used simulated experiments to demonstrate that our frequency-normalized resampling algorithm recovers the ground truth patch importance. We simulated two classes of images as shown in Fig. \ref{synth}, with background = 0 and strips = 1 and Gaussian noise ($\mu = 0,\sigma = 0.01$). They can be gridded into 9 patches. We assumed that patch B of class 0 and 1 are $\bf{equally\ important}$ to human understanding. However, in our synthetic model, the sample set was biased with 900 images in class 0 and 100 images in class 1. A simple 2-layer convolutional neural network was used as the image classifier, which achieved 100\% classification accuracy. Since the shift of corrupted images was obvious, we used misclassification rate to measure whether $p(c|\bm{X_{\setminus r}})$ was approximated reasonably by equally weighted sampling (which means $w_i = 1/K$) or by our frequency-normalized sampling. In Table 1,  our frequency-normalized sampling approach ('Normalize') is superior to the equally weighted one ('Equal') in treating patch B equally in both classes.

%\begin{figure}[t]
%	\centering
%	\centerline{\includegraphics[width=11cm]{exp1}}
%	%\centerline{(a) Result 1}\medskip
%	\label{exp1}
%		%
%\end{figure}
\subsection{Experiment 2: Task-fMRI Experiment}
We tested our methods on a group of 82 ASD children and 48 age and IQ-matched healthy controls. Each subject underwent a task fMRI scan (BOLD, TR = 2000ms, TE = 25ms, flip angle = $60^{\circ}$, voxel size $3.44\times3.44\times4 mm^3$) acquired on a Siemens MAGNETOM Trio TIM 3T scanner.

For the fMRI scans, subjects performed the "biopoint" task, viewed point light animations of coherent and scrambled biological motion in a block design \cite{Kaiser07122010} ($24s$ per block). The fMRI data was preprocessed using FSL \cite{smith2004advances} for 1) motion correction, 2) interleaved slice timing correction, 3) BET brain extraction, 4) spatial smoothing (FWHM=5mm), and 5) high-pass temporal filtering. The functional and
anatomical data were registered and parcellated by AAL atlas \cite{tzourio2002automated} resulting in 116 ROIs. %\cite{venkataraman2016bayesian}.  
We applied a sliding window ($w=3$) along the time dimension of the 4D fMRI, generating 144 3D volume pairs (mean and std) for each subject.
 
\begin{figure}[t]
	\centering
	\centerline{\includegraphics[width=12cm]{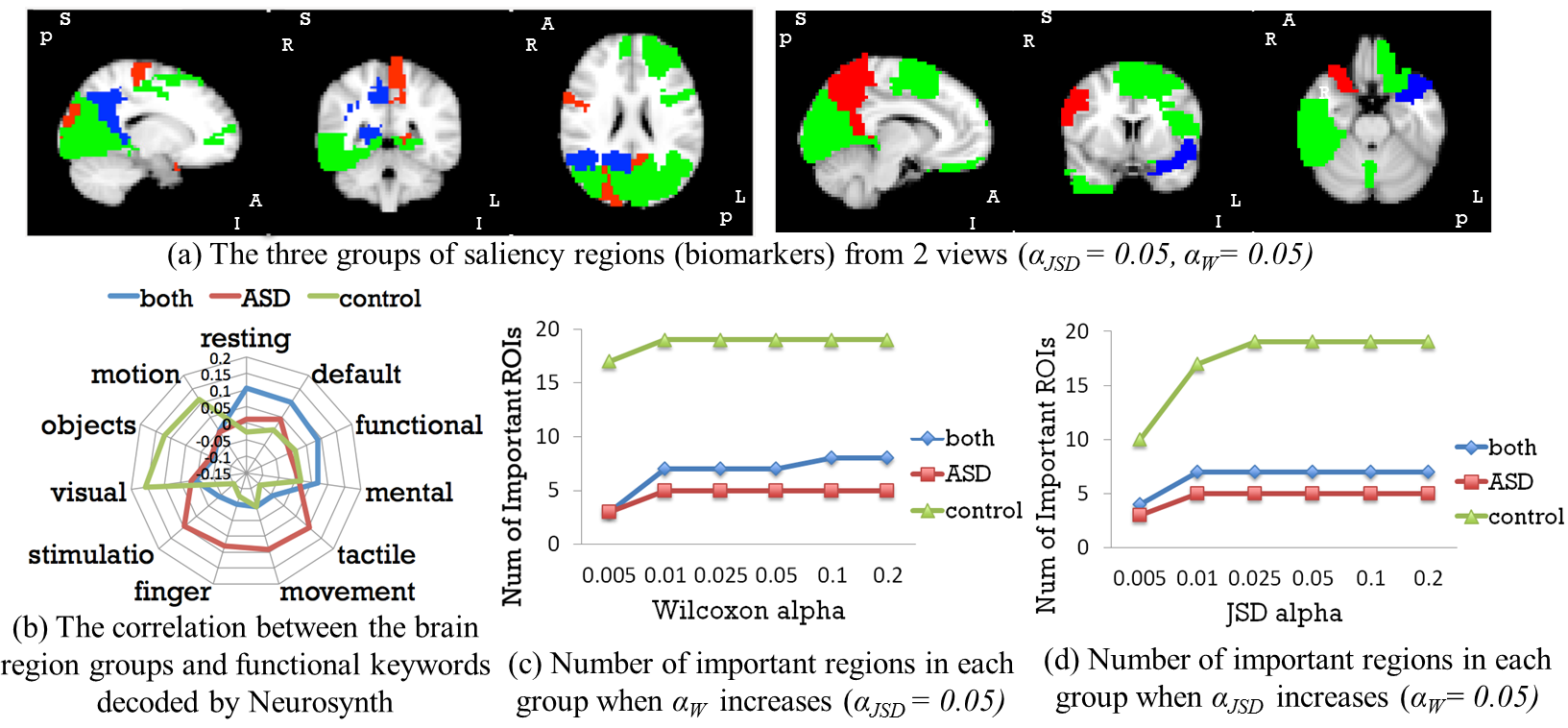}}
	%  \vspace{2.0cm} %used to be biopoint
	%\centerline{(a) Result 1}\medskip

	\caption{Important Biomarkers Detected in Biopoint Dataset}
	\label{biopoint}
	%\vspace{-.5cm}
\end{figure}

\begin{figure}[t]
	\centering
	\centerline{\includegraphics[width=12cm]{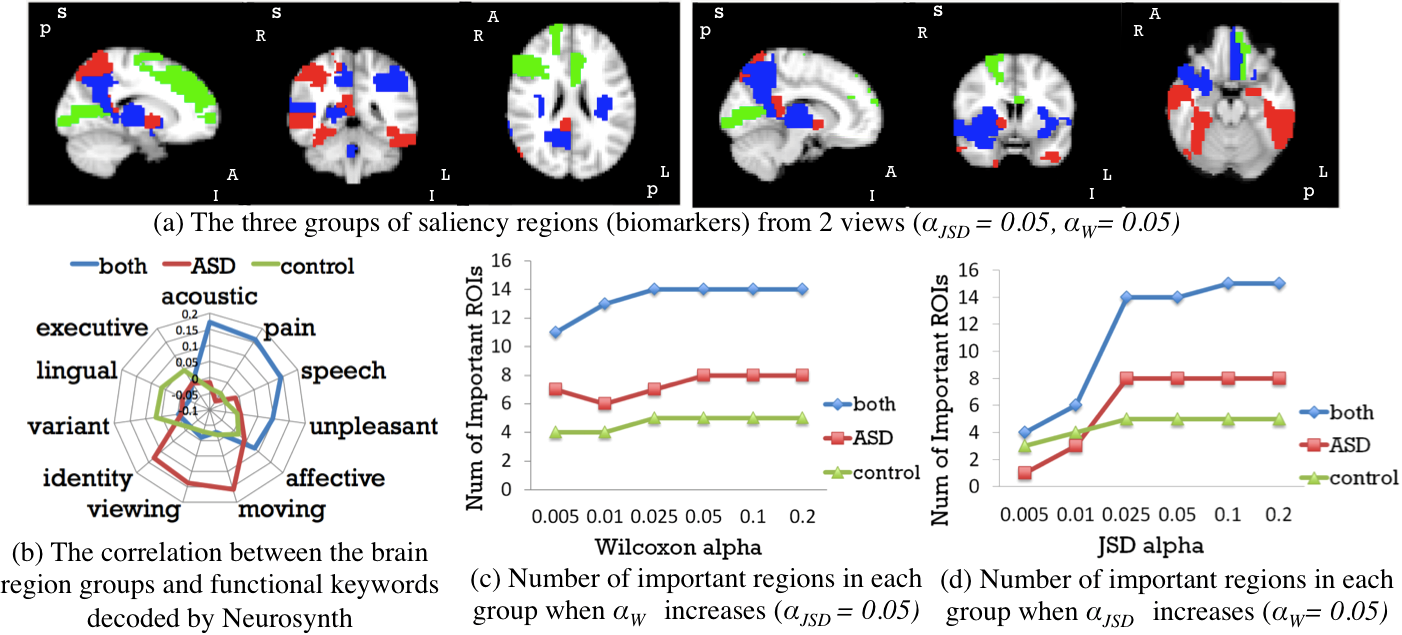}}
	%  \vspace{2.0cm} % use to be ABIDE
	%\centerline{(a) Result 1}\medskip
	
	\caption{Important Biomarkers Detected in ABIDE Dataset}
	\label{abide}
	%\vspace{-.5cm}	%
\end{figure}
We split 85\% subjects (around $16k$ 3D volume pairs) as training set, 7\% as validation set for early stopping and 8\% as testing set, stratified by class. The model achieved 87.1\% accuracy when evaluated on each 3D pair input of the testing set. Fig. \ref{biopoint} (a) and (b) give two views of the important ROIs brain map ($\alpha_{JSD} = 0.05$, $\alpha_{W}=0.05$). Blue ROIs are associated with identifying both ASD and control. Red ROIs are associated with identifying ASD only and green ROIs are associated with identifying control only. By decoding the neurological functions of the important ROIs with Neurosynth \cite{yarkoni2011large}, we found 1) regions related to default mode and functional connectivity  are significant in classifying both individuals with ASD and controls, which is consistent with prior literature related to executive functioning and problem-solving in ASD \cite{Kaiser07122010}; 2) regions associated with finger movement are relevant in classifying individuals with ASD, and 3) visual regions were involved in classifying controls, perhaps because controls may attend to the visual features more closely, whereas ASD subjects tend to count the dots on the video \cite{ventola2007differentiating}. 

%\vspace{-.5cm}
\begin{figure}[t]
	\centering
	\centerline{\includegraphics[width=11cm]{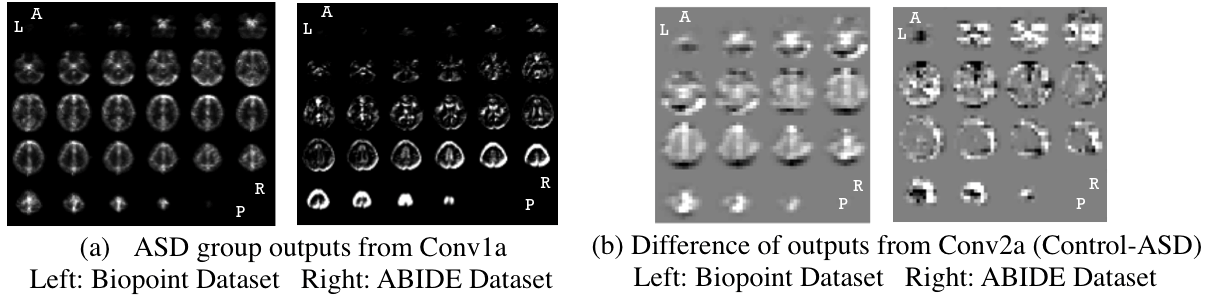}}
	%  \vspace{2.0cm}
	%\centerline{(a) Result 1}\medskip
	%\vspace{-.7em}
	\caption{Intermediate outputs (activation maps) of DNN}
	\label{middel}
	%\vspace{-.7cm}	%
\end{figure}

\subsection{Experiment 3: Resting-state fMRI}
%\vspace{-.25cm}
 
We also performed experiments on data from the ABIDE I cohort UM site
\cite{di2014autism,smith2004advances}.This resulted in 41 ASD subjects and 54
healthy controls. Each subject initially had 293 frames. As in the task-fMRI experiment, we generated 2-channel images. We used the weights of the pre-trained 2CC3D networks in experiment 2 as our initial network weights. We split 33 ASD subjects and  43 controls for training (around 22k 3D volume pairs). 9 subjects were used as validation data for early stopping. The classifier achieved $85.3\%$ accuracy in identifying individual 3D volume on the 10 subjects testing set. The biomarker detection results are shown in Fig. \ref{abide}: 1) emotion related regions colored in blue are highlighted for both groups; 2) regions colored in red (viewing and moving related) are associated with identifying ASD; and 3) green regions (related to executive and lingual) are associated with identifying control.  

\subsection{Results Analysis}
%\vspace{-.25cm}
In experiment 2, since the subjects were under visual task, visual patterns were detected. Whereas in experiment 3, subjects were in resting state, so no visual regions were detected. In addition, we found many common ROIs in both experiments: frontal (motivation related), precuneus (execution related), etc. Previous research \cite{Kaiser07122010} also indicated these regions are associated with identifying ASD vs. control. Moreover, from the sub-figure (c), (d) of Fig. \ref{biopoint} and \ref{abide}, the groups of detected important regions are very stable when tuning JSD confidence level (1-$\alpha_{JSD}$) and Wilcoxon testing threshold $\alpha_W$, except when $\alpha_{JSD}$ is very small. This is likely because the original prediction distribution is fat tailed. Furthermore, we validate the results with the activation maps from the 1st and 2nd layers of the DNN. The output of each filter was averaged for 10 controls and for 10 ASD subjects. The 1st convolutional layer captured structural information and distinguished gray vs. white matter (Fig. \ref{middel}(a)). Its outputs are similar in both control and ASD group. The outputs of the 2nd convolutional layer showed significant differences between groups in Fig. \ref{middel}(b). Regions darkened and highlighted in Fig. \ref{middel}(b) correspond to many regions detected by our proposed method.

\section{Conclusions} 
%\vspace{-.2cm}
We designed a 2-stage (DNN + prediction distribution analysis) pipeline to detect brain region saliency for identifying ASD and control subjects.  Our sampling and significance testing scheme along with the accurate DNN classifier ensure reliable biomarker detection results. Our method was designed for interpreting important ROIs for registered images, since the traditional machine learning feature selection methods can not be directly used in interpreting DNNs. Moreover, our proposed method can be directly used to interpret any other machine learning classifiers.  Overall, the proposed method provides an efficient and objective way of interpreting the deep learning model applied to neuro-images.
%\vspace{-.2cm}

\bibliographystyle{ieeetr}
\bibliography{MICCAI}

\end{document}